\documentclass{article}

\usepackage{microtype}
\usepackage{graphicx}
\usepackage{subfigure}
\usepackage{booktabs} % for professional tables

\usepackage{hyperref}
\usepackage[accepted]{icml2024}
\usepackage{amsmath}
\usepackage{amssymb}
\usepackage{mathtools}
\usepackage{amsthm}
\usepackage[capitalize,noabbrev]{cleveref}

\theoremstyle{plain}

\theoremstyle{definition}

\theoremstyle{remark}

\usepackage[textsize=tiny]{todonotes}

\icmltitlerunning{Community search signatures as foundation features for human-centered geospatial modeling}

\begin{document}

\twocolumn[
\icmltitle{Community search signatures as foundation features for human-centered geospatial modeling}

\icmlsetsymbol{equal}{*}

\begin{icmlauthorlist}
\icmlauthor{Mimi Sun}{comp}
\icmlauthor{Chaitanya Kamath}{comp}
\icmlauthor{Mohit Agarwal}{comp}
\icmlauthor{Arbaaz Muslim}{comp}
\icmlauthor{Hector Yee}{comp}
\icmlauthor{David Schottlander}{comp}
\icmlauthor{Shailesh Bavadekar}{comp}
\icmlauthor{Niv Efron}{comp}
\icmlauthor{Shravya Shetty}{comp}
\icmlauthor{Gautam Prasad}{comp}
\end{icmlauthorlist}

\icmlaffiliation{comp}{Google Research}

\icmlcorrespondingauthor{Mimi Sun}{\hbox{mimisun@google.com}}
\icmlcorrespondingauthor{Gautam Prasad}{\mbox{gautamprasad@google.com}}

\icmlkeywords{Machine Learning, ICML, datasets}

\vskip 0.3in
]

\printAffiliationsAndNotice{}  % leave blank if no need to mention equal contribution

\begin{abstract}
Aggregated relative search frequencies offer a unique composite signal reflecting people's habits, concerns, interests, intents, and general information needs, which are not found in other readily available datasets.  Temporal search trends have been successfully used in time series modeling across a variety of domains such as infectious diseases, unemployment rates, and retail sales.  However, most existing applications require curating specialized datasets of individual keywords, queries, or query clusters, and the search data need to be temporally aligned with the outcome variable of interest.  We propose a novel approach for generating an aggregated and anonymized representation of search interest as foundation features at the community level  for geospatial modeling.  We benchmark these features using spatial datasets across multiple domains.  In zip codes with a population greater than 3000 that cover over 95$\%$ of the contiguous US population, our models for predicting missing values in a 20$\%$ set of holdout counties achieve an average R\textsuperscript{2} score of 0.74 across 21 health variables, and 0.80 across 6 demographic and environmental variables. Our results demonstrate that these search features can be used for spatial predictions without strict temporal alignment, and that the resulting models outperform spatial interpolation and state of the art methods using satellite imagery features.  
\end{abstract}

\section{Introduction}
\label{introduction}

Geospatial modeling is an important tool for generating actionable maps of population dynamics \cite{hay2004global,mellor2023understanding} and environmental conditions \cite{lam2023learning,nearing2024global}. It is particularly useful in situations involving either limited scale and resolution (such as surveys like the US census) or complex combinations of a multitude of factors (such as human health, which involves environmental and genetic factors). We focus on three key spatial modeling tasks:

\begin{itemize}
\item \textbf{Imputation.} Missing data is a common issue with surveys such as the census that might miss certain demographics, locations, or answers \cite{brick1996handling}. The goal of imputation is to generate the most likely data to substitute those missing values. A model built for imputation can learn from locations where a value is known, previous time points of a value, or the connection to related variables.
\item \textbf{Extrapolation.} Data may be missing from large contiguous areas, for example measures such as socioeconomic data may be missing group countries where traditional  inventories aren't available \cite{steele2021mobility}. A model trained on ground truth data collected from some countries can be used to make inferences about other countries based on underlying global signals such as those from remote sensing, search signatures, mobility, or cell phone usage data.
\item \textbf{Super-Resolution.} Small area estimation or spatial disaggregation is necessary to get data at finer spatial scales for better targeting and serving local communities. For example, disaggregating census data has been shown as a way to model populations at a high resolution \cite{stevens2015disaggregating}. In practice, this could mean that a model that is trained on ground truth data at the county level could be used to make inferences at the zip code level.
\end{itemize}

Web search queries have been demonstrated to be a powerful signal for geospatial forecasting \cite{ginsberg2009detecting,choi2012predicting} and range of specific use cases ranging from economic to epidemiological \cite{jun2018ten}. Our focus here is to evaluate a general representation of web search queries that can be used as a foundation feature, applicable to a wide range of spatial modeling tasks with relatively simple downstream models. Previous work using web search queries has centered on hand crafted features for specific use cases. However, keyword and query curation is time consuming and requires domain knowledge. In addition, many keywords of interest are rarely searched and result in spotty spatial coverage.

Billions of search queries are made each day across the US. However, when it comes to smaller regions, such as zip codes with less than 10000 people, and specific time ranges, July 2020, the search data on more niche topics (such as "heat stroke" and "anosmia") is much more sparse. As a result, feature sets created from those keywords are relatively noisy and have lower spatial and temporal coverage \cite{bavadekar2020google}. On the other hand, top searches such as "youtube" and "weather" have sufficient volume in most zip codes. For each geographic region, we can create a relatively low dimensional feature vector (local search signature) from its top searches. These vectors can then be used to train models for a variety of downstream tasks in all locations. Label super-resolution is possible because the vectors reside in the same space across all geographic granularities. For example, a model trained on search signatures and ground truth data about disease prevalence at the county level can be used to make inferences from search signatures at the zip code level.

We design a general purpose representation of web search queries and benchmark across a range of geospatial tasks including imputation, extrapolation, and super-resolution.

Our contributions include the following.
\begin{itemize}
\item We showcase a novel general purpose representation of web search data that can be used as foundation features without keyword curation, timeseries, or matching to specific time points of target variables.
\item We demonstrate that these foundation search features encode enough information for training downstream models in multiple human-centered domains to achieve state of the art performance: health, income, population density, housing, environment.
\end{itemize}

\section{Related Work}
Search engine queries have been analyzed to track various symptoms, disease prevalence and economic variables. In \cite{ginsberg2009detecting, santillana2015combining, paul2014twitter}, researchers were able to nowcast and forecast influenza outbreaks leveraging the high-correlation of certain search queries and tweets with influenza-like symptoms.  Similarly, in \cite{choi2012predicting}, authors use search frequencies to predict the “present” socio-economic variables, which is extended to forecasting and fine-grained spatial resolution in \cite{wu2015future} for housing prices. Such methods establish the prediction strength of using search features as proxy for any geographical surveillance, often taking a span of weeks, months or even years in some cases.

Complementing the insights gleaned from search data, significant research has focused on leveraging visual features in satellite imagery, and other geotagged metadata for downstream geospatial prediction tasks. GPS2Vec \cite{yin2019gps2vec} learns semantic embeddings for geo-coordinates combining with geotagged documents including tweets, images, check-ins etc. For instance, MOSAIKS \cite{rolf2021generalizable} combine the satellite imagery with ML to obtain a single encoding for diverse prediction tasks including housing prices, road length, forest cover, etc. CSP \cite{mai2023csp} proposed contrastive self-supervised learning with images for FMoW \cite{christie2018functional} and iNAT datasets \cite{van2018inaturalist}. SatCLIP \cite{klemmer2023satclip} applies the contrastive learning to worldwide available S2 satellite imagery showcasing superior performance on multiple regression and classification tasks.

\section{Methods}

We focus our analysis at the US zip code level because this is the smallest spatial division commonly used by both the public and private sectors, as well as consumers. To simplify the analysis, we treat zip codes and Zip Code Tabulation Areas (ZCTAs) from the US Census Bureau as equivalent. This is a common practice for large scale analysis at the state and national levels. Crosswalk files and methods are available for more precise joining of zip codes and ZCTAs when needed but they are outside the scope of this study.  

The study area is defined as the contiguous United States (CONUS), which includes 48 states and the District of Columbia.

\subsection{Search Feature Dataset Creation}

For each zip code in the US, we extract the top 500 queries that were searched for at least 20 times in that zip code over the month of July 2022. This yielded over 1MM unique queries. We then ranked the queries by universal popularity, quantified by the total number of zip codes that they appeared in. The list of top 1000 queries are kept as the ranked vocabulary and the query text is replaced with numerical feature IDs. The least common query included in the vocabulary appeared in over 700 zip codes. For each zip code, we then created a 1000 dimensional feature vector from in-vocab query counts, and scaled the values so that they sum to 100. For the purposes of this study, we used these 1000 dimensional vectors directly as local search signatures.  All queries included in the subsequent analysis occurred at least 14000 times in the dataset. 

\begin{figure}[h!]
%\vskip 0.2in
% \begin{center}
\subfigure[]{\includegraphics[width=0.72\columnwidth]{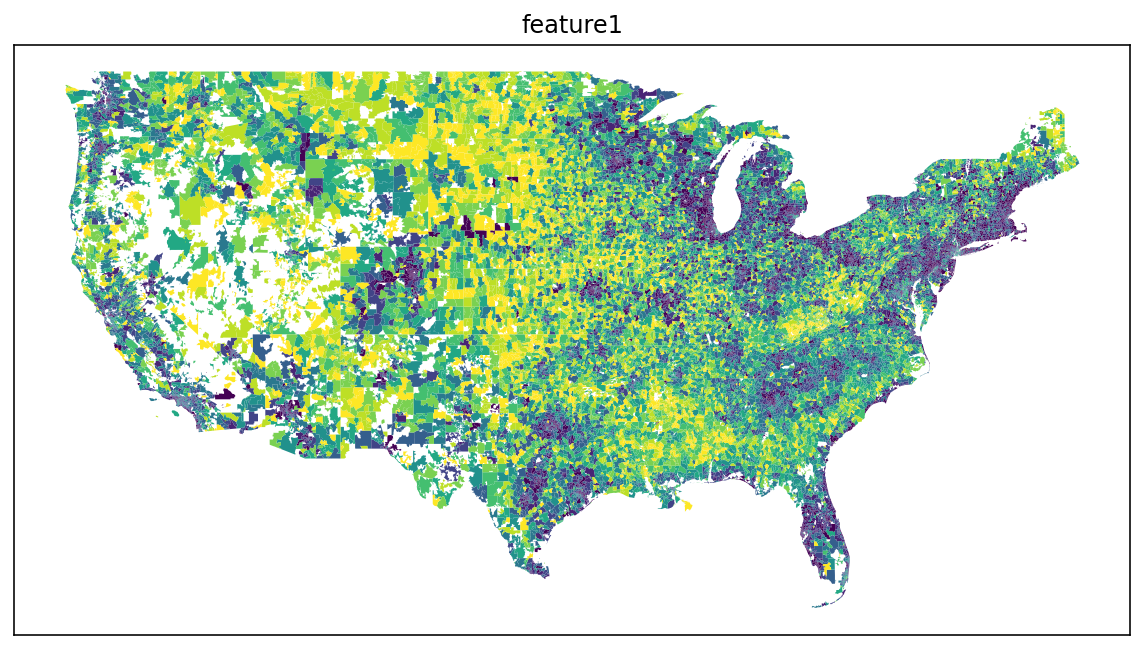}}
\subfigure[]{\includegraphics[width=0.27\columnwidth]{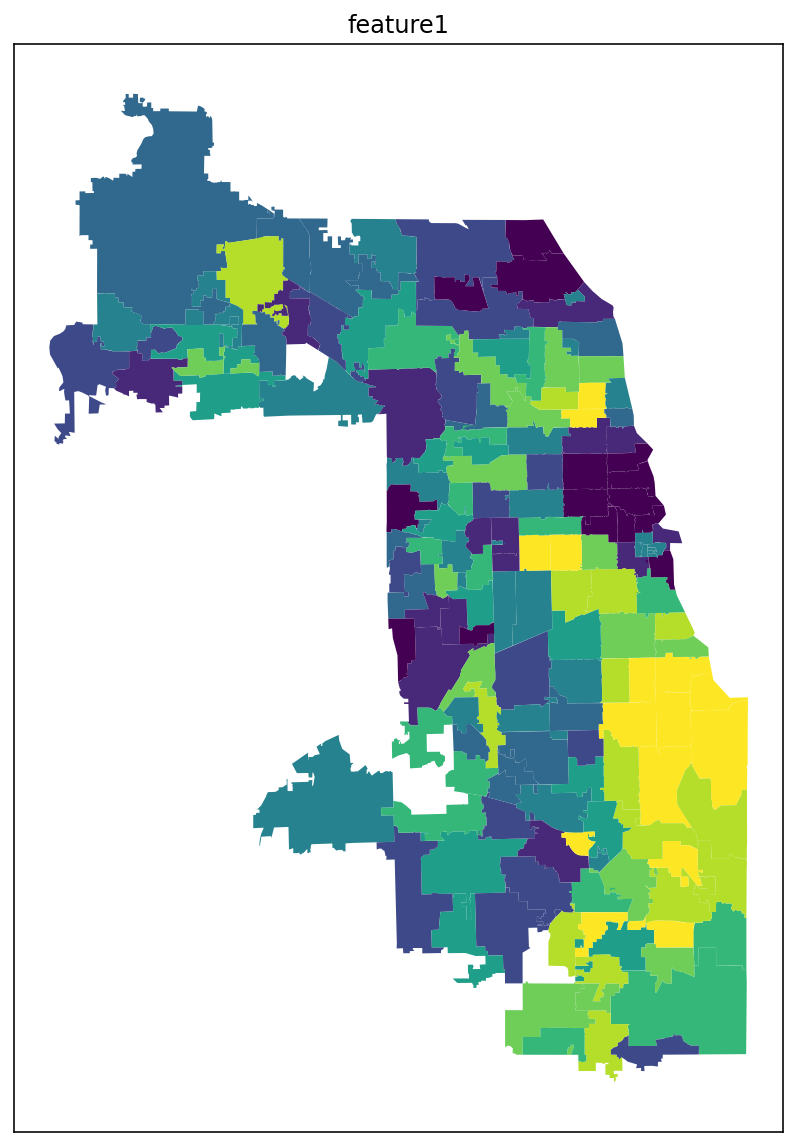}}
\subfigure[]{\includegraphics[width=0.72\columnwidth]{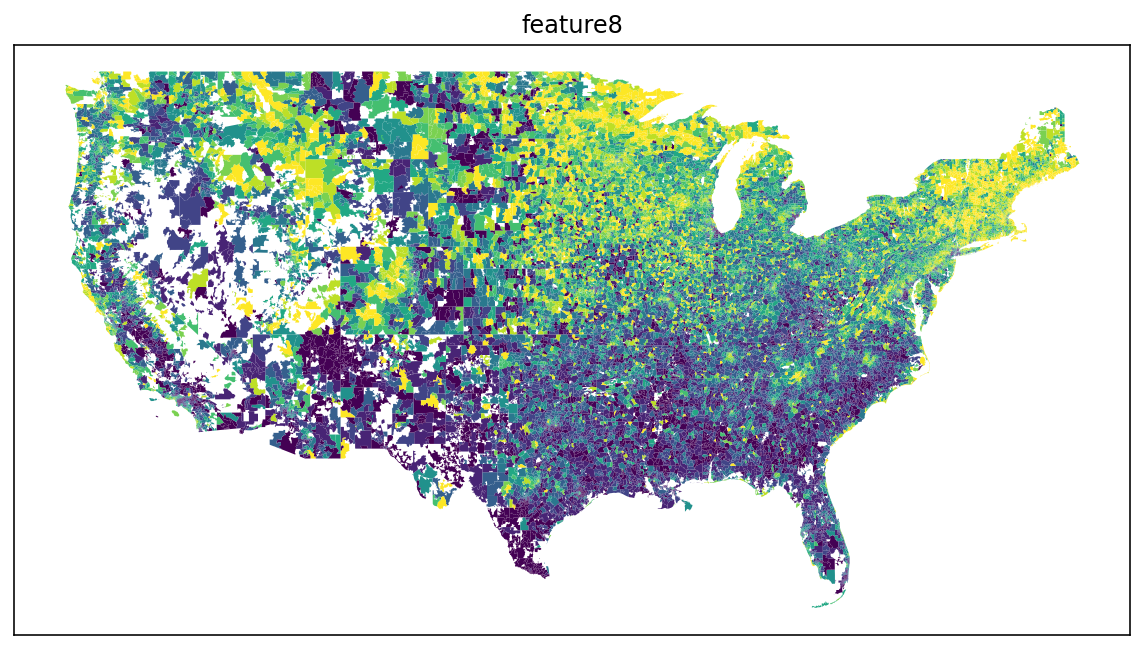}}
\subfigure[]{\includegraphics[width=0.27\columnwidth]{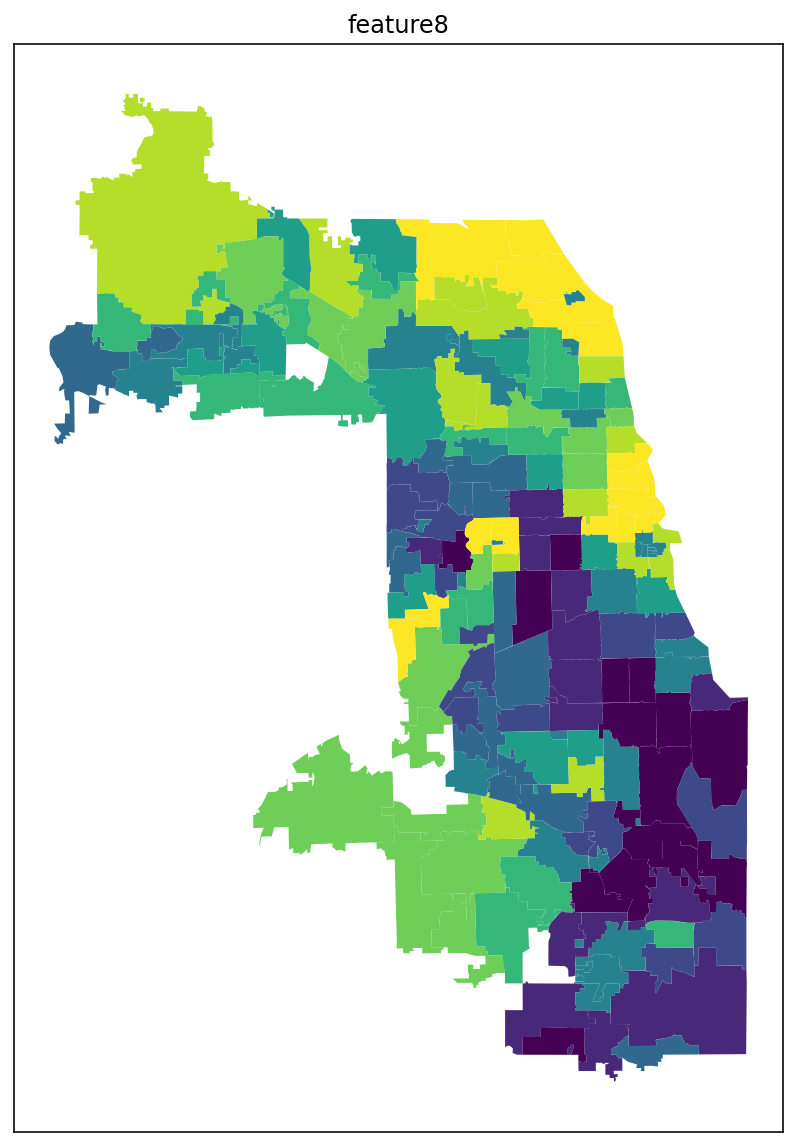}}
\caption{First row shows the values of "feature1" in CONUS and Cook County, IL. The second row shows "feature8" in the same locations.}
\label{feature_maps}
% \end{center}
\vskip -0.2in
\end{figure}

Search data is not logged for regions with low population and regions that are under 3 square-kilometers in area. In addition, we also filtered out zip codes where the feature vector contained over 98$\%$ 0s. Overall approximately 28K zip codes are present in the dataset, covering $>$ 95$\%$ of the CONUS population. For the remaining zip codes, we estimate the search signatures by filling in the median value of the parent county. Figure \ref{feature_maps} maps two sample features at the national and county levels to illustrate the spatial resolution, coverage, and variations for the dataset. 

We also compared search feature datasets from different time periods which yielded equivalent results in model performance.

\subsection{Benchmark Construction}
\label{sec:benchmark}

We include all variables (or equivalents) modeled in the MOSAIKS paper \cite{rolf2021generalizable} except for road lengths. In addition, we modeled 21 health-related variables from the CDC PLACES dataset \cite{cdcplaces2024}. 

All benchmark data are publicly available through either Data Commons \cite{datacommons2024} or the Google Earth Engine Catalog \cite{gorelick2017google}. 

Raster image data were aggregated to the zip code level using the 2010 ZCTA boundaries \cite{zcta2010}. The original sources of the data varied by variable. Tree cover data was obtained from ESA WorldCover \cite{zanaga2022esa}, elevation from NASA SRTM Digital Elevation \cite{farr2007shuttle}, and nighttime lights from VIIRS \cite{elvidge2021annual}.

The tabular data from Data Commons are provided for Zip Code Tabulation Areas (ZCTAs) which are not always equivalent to zip codes, however they are generally treated as equivalent for large scale analysis at the state and national levels. Crosswalk files and methods are available for more precise joining of zip codes and ZCTAs when needed but they are outside the scope of this study.

\subsection{Holdout Test Set Construction}

\begin{figure}[ht!]
\vskip 0.2in
\begin{center}
\centerline{\includegraphics[width=\columnwidth]{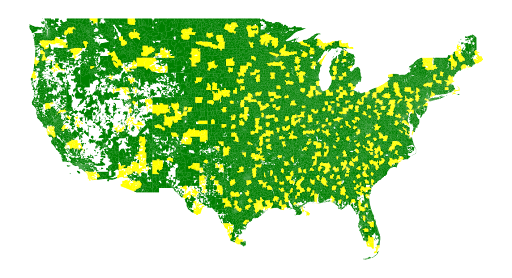}}
\caption{Map showing the dataset split. Yellow areas are counties where all contained zip codes are in the holdout set, green areas are used for training and validation.}
\label{data_split}
\end{center}
\vskip -0.2in
\end{figure}

\begin{figure*}[h!]
\vskip 0.2in
\begin{center}
\includegraphics[width=12cm]{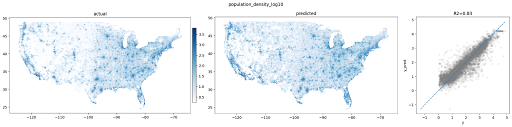}
\includegraphics[width=12cm]{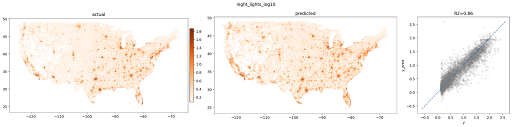}
\includegraphics[width=12cm]{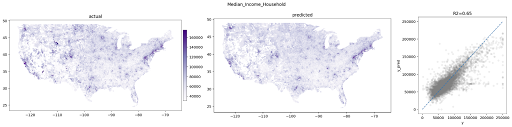}
\includegraphics[width=12cm]{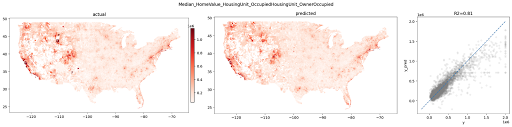}
\includegraphics[width=12cm]{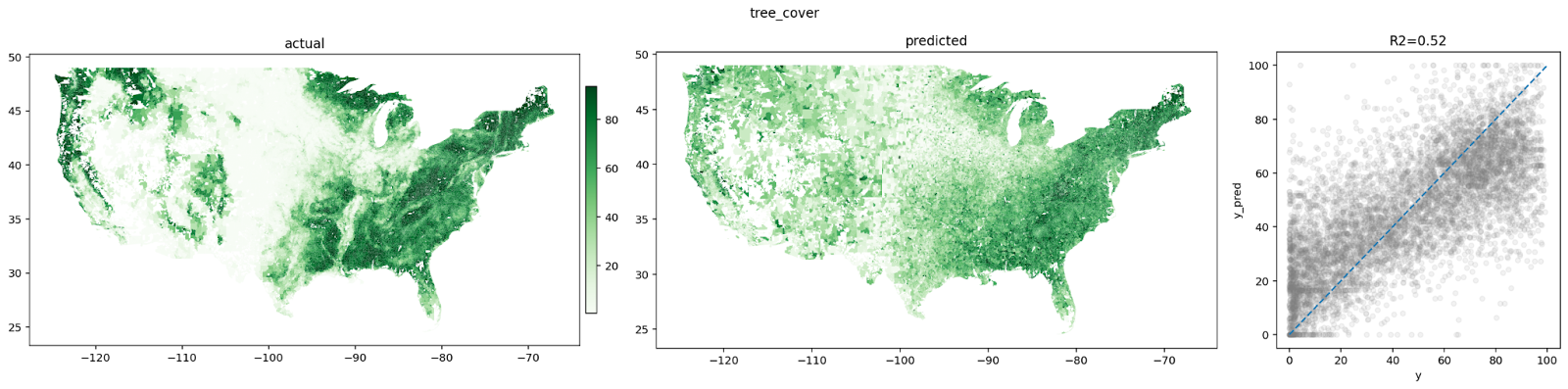}
\includegraphics[width=12cm]{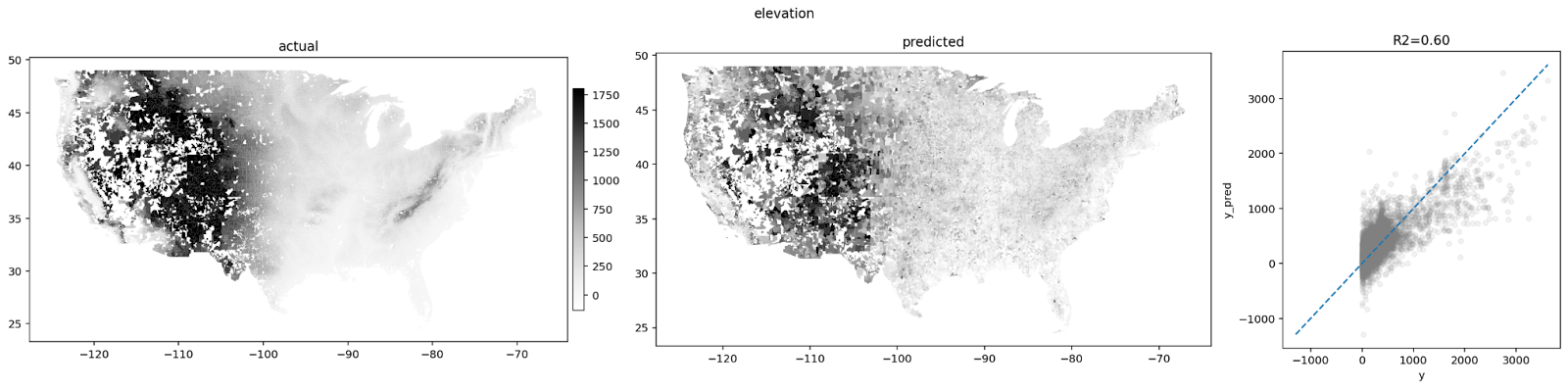}
\caption{Actual, predicted, and test set scatter plot of 6 variables. The "predicted" column shows a concatenation of predictions for the 5 validation sets and the test set, made by six different models. The scatter plot shows the test set performance. Training set predictions are not displayed.}
\label{imputation_maps}
\end{center}
\vskip -0.2in
\end{figure*}

First, we created a one-to-one mapping from zip codes to counties based on the largest land area overlap. Then we randomly selected 20$\%$ of the counties and all of the zip codes contained within them as the holdout test set.  This approach creates more spatial separation between the training and test data than uniformly selecting zip codes for the test set, thereby mimicking the more difficult types of real world missing-data situations more closely.  Figure \ref{data_split} shows the different areas that are used for training and testing. 

We used 5-fold cross-validation on the training dataset during model development. The holdout set was only used for final testing and visualization.

\subsection{Modeling and evaluation}
To demonstrate the predictive power and ease of use of the search features, we fit a simple linear model using Ridge regression for each prediction task. The regularization term is lightly tuned through cross validation, and each model runs in a few seconds on a single CPU. 

For comparison, we implemented two commonly used methods for predicting spatial missing values: Inverse Distance Weighting (IDW) for spatial interpolation and Hierarchical Median Imputation for structured missing value imputation. Both methods are chosen for their comparable ease of use and reliable performance without careful tuning.  For spatial interpolation, we used the geographic centroids of the zip codes for all prediction tasks. Switching to using population weighted centroids may improve the performance on some tasks \cite{henry2008estimating}.  For imputation, we use the parent county median label value where available, and the state level median otherwise.  

For evaluation, we measured the coefficient of determination (R\textsuperscript{2}) of all experiments and report them in the next section.

\section{Experiments}

\subsection{Imputation Tasks}

We benchmarked the search features using 5-fold cross validation and visualized the results in Table \ref{imputation_table}.  Our models performed better than MOSAIKS on 4 out of the 6 tasks, with an average R\textsuperscript{2} of 0.71 compared to 0.68.  In another experiment we limited our training and test set to zip codes with a population greater than 3000, which is similar to the population-weighted sampling used for some variables in MOSAIKS. Our reduced dataset still covers over 95$\%$ of the US population, and the test set R\textsuperscript{2} average has increased to 0.80.

\begin{table}[h!]
\caption{Test set R\textsuperscript{2} metric comparison across different tasks and models. The MOSAIKS values are from \cite{rolf2021generalizable}. All other results are obtained  using the same holdout data.}
\label{imputation_table}
\vskip 0.15in
\begin{center}
\begin{small}
\begin{sc}
\resizebox{\columnwidth}{!}{%
\begin{tabular}{lcccc}
\toprule
Task & IDW & MOSAIKS & TopSearch & \begin{tabular}{@{}c@{}}TopSearch \\ -pop3k\end{tabular} \\
\midrule
Median Household Income & 0.35 & 0.45 & 0.65 & \textbf{0.79} \\
Housing Price, Home Value & 0.66 & 0.46 & 0.81 & \textbf{0.84} \\
Night lights & 0.57 & 0.85 & 0.86 & \textbf{0.89} \\
Population Density & 0.64 & 0.73 & 0.83 & \textbf{0.90} \\
Forest Cover & 0.70 & \textbf{0.91} & 0.51 & 0.63 \\
Elevation & \textbf{0.94} & 0.68 & 0.60 & 0.74 \\
\textbf{Mean} & 0.64 & 0.68 & 0.71 & \textbf{0.80} \\
\bottomrule
\end{tabular}
}
\end{sc}
\end{small}
\end{center}
\vskip -0.1in
\end{table}

As further illustrated in Figure \ref{imputation_maps}, our models using the search features perform very well on human-centered variables, while satellite imagery features excelled at predicting forest cover amount. This demonstrates that these two feature sets could be complementary to each other in geospatial modeling, and a true geospatial foundation model should make use of both types of features.

\begin{figure}[h!]
\vskip 0.2in
\begin{center}
\includegraphics[width=5cm]{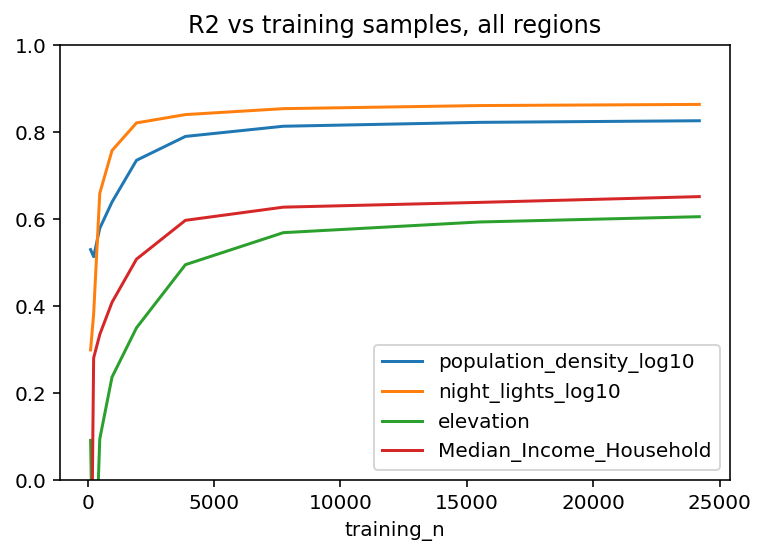}
\includegraphics[width=5cm]{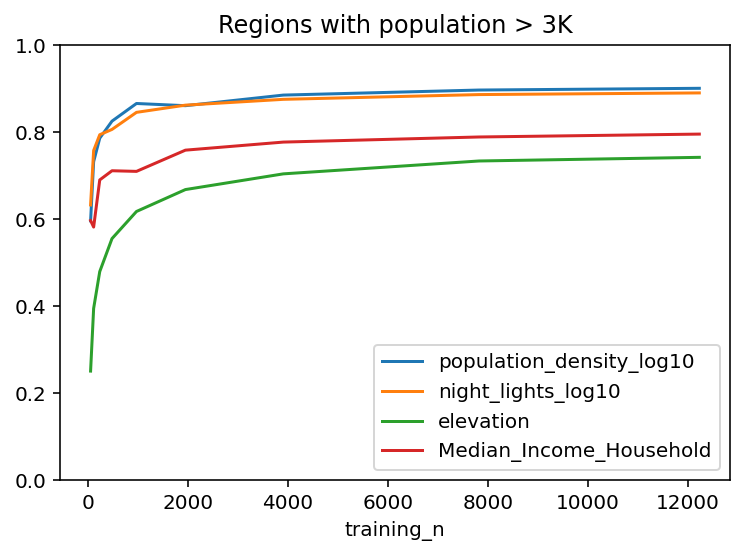}
\caption{Performance vs training data \%}
\label{performance_training}
\end{center}
\vskip -0.2in
\end{figure}

\begin{figure}[h!]
\vskip 0.2in
\begin{center}
\includegraphics[width=5cm]{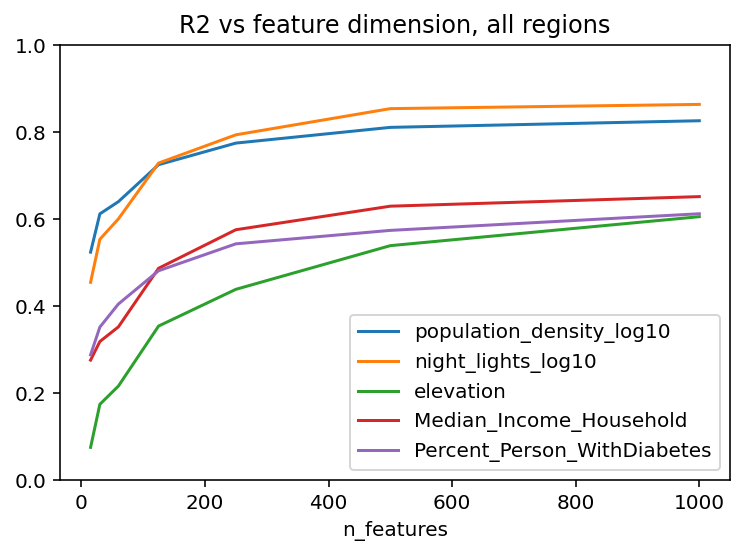}
\includegraphics[width=5cm]{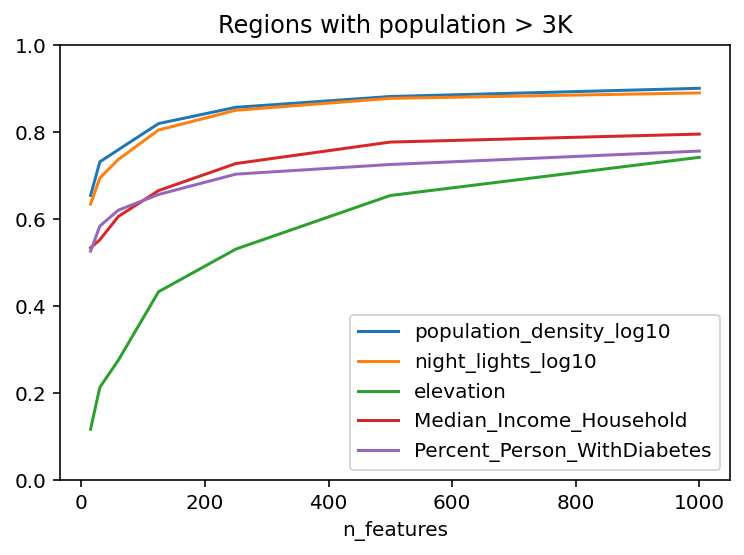}
\caption{Performance vs feature dimension}
\label{performance_feature}
\end{center}
\vskip -0.2in
\end{figure}

We also experimented with varying training sample sizes and feature dimensions. The model performance generally remained good for as few as 5000 train samples and as few as 500 features. Figures \ref{performance_training} and \ref{performance_feature} show the variations for a few of the prediction tasks.

\begin{figure*}[h!]
%\vskip 0.2in
\begin{center}
\includegraphics[width=12cm]{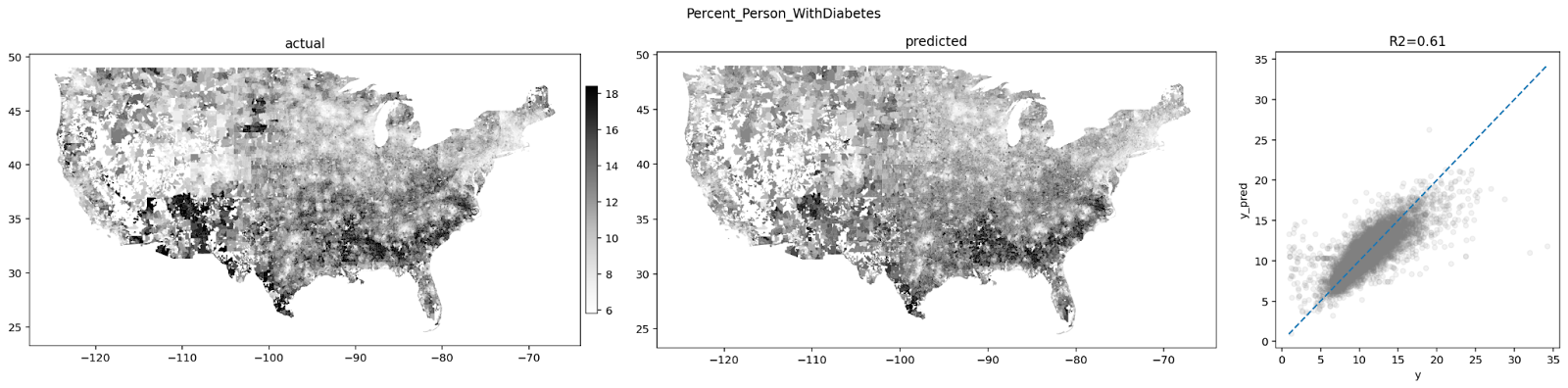}
\includegraphics[width=12cm]{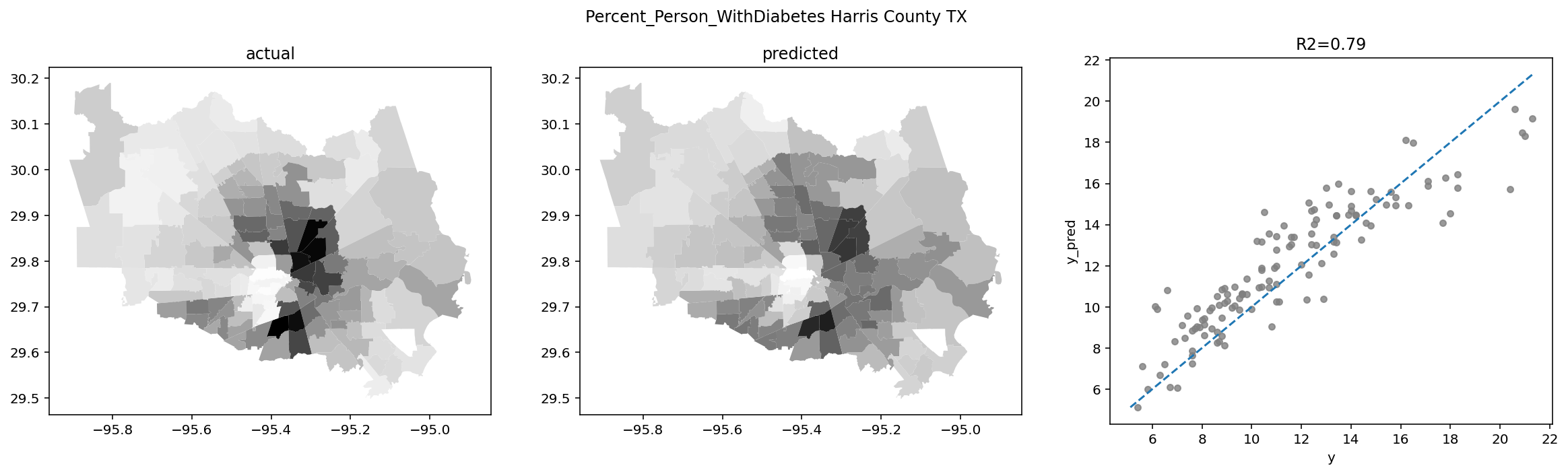}
\caption{The predicted vs. actual values of diabetes prevalence, nationally and in Harris County, TX from the holdout set, which is the most populous county in that set.}
\label{diabetes_imputation}
\end{center}
%\vskip -0.2in
\end{figure*}

In addition to the demographic and environmental variables above, we also modeled 21 health-related variables from the CDC PLACES dataset \cite{cdcplaces2024} in Table \ref{health_table}. Our model obtained an average R\textsuperscript{2} value of 0.61 in all zip codes, and 0.74 in zip codes with a population over 3000. As shown in Figure \ref{diabetes_imputation}, qualitatively, the model is able to make good predictions for large holdout counties. We would like to carry out a more quantitative analysis of within county performance in a future study.

\begin{table}[h!]
\caption{Holdout set R\textsuperscript{2} for all 21 health indicator variables from CDC PLACES}
\label{health_table}
\vskip 0.1in
\begin{center}
\begin{small}
\begin{sc}
\resizebox{\columnwidth}{!}{%
\begin{tabular}{lccc}
\toprule
Health Variable & IDW & TopSearch & \begin{tabular}{@{}c@{}}TopSearch \\ -pop3k\end{tabular} \\
\midrule
High Cholesterol & 0.23 & 0.46 & \textbf{0.61} \\
Physical Health Not Good & 0.42 & 0.68 & \textbf{0.81} \\
Stroke & 0.28 & 0.57 & \textbf{0.75} \\
Binge Drinking & 0.39 & 0.46 & \textbf{0.57} \\
Physical Inactivity & 0.42 & 0.72 & \textbf{0.82} \\
Received Annual Checkup & 0.67 & 0.72 & \textbf{0.82} \\
Cancer (excl. Skin Cancer) & 0.24 & 0.54 & \textbf{0.71} \\
Diabetes & 0.32 & 0.61 & \textbf{0.76} \\ 
Mental Health Not Good & 0.29 & 0.60 & \textbf{0.74} \\
Coronary Heart Disease & 0.40 & 0.56 & \textbf{0.73} \\
High Blood Pressure & 0.45 & 0.64 & \textbf{0.78} \\
Received Cholesterol Screening & 0.36 & 0.50 & \textbf{0.68} \\
Received Dental Visit & 0.38 & 0.74 & \textbf{0.83} \\
Asthma & 0.38 & 0.60 & \textbf{0.75} \\
Chronic Kidney Disease & 0.24 & 0.49 & \textbf{0.66} \\
Arthritis & 0.45 & 0.64 & \textbf{0.78} \\
COPD & 0.50 & 0.67 & \textbf{0.82} \\
High Blood Pressure (Medicated) & 0.16 & 0.40 & \textbf{0.59} \\
Obesity & 0.48 & 0.74 & \textbf{0.80} \\
Sleep Less Than 7 Hours & 0.39 & 0.68 & \textbf{0.75} \\
Smoking & 0.44 & 0.72 & \textbf{0.84} \\
\textbf{Mean} & 0.38 & 0.61 & \textbf{0.74} \\
\bottomrule
\end{tabular}
}
\end{sc}
\end{small}
\end{center}
\vskip -0.1in
\end{table}

\subsection{Extrapolation and Large Area Imputation Tasks}

Sometimes data is missing from entire states in a national dataset, and some other datasets are only available from a few states' local sources while missing from the rest of the country. We conducted some experiments to evaluate our search features' ability to generalize across states in these scenarios. 

To test the scenario where a few states are missing, we split the training dataset by the 48 CONUS states and D.C., and performed 10-fold cross validation where each model was trained on data from 44 or 45 regions and tested on the remaining 5 or 4 missing regions. The search features model still performs relatively well on the majority of tasks except for tree cover and elevation. Excluding those two tasks, the average R\textsuperscript{2} of TopSearch across all 25 tasks is 0.40, and 0.54 for regions with populations over 3000.

\begin{figure*}[h!]
\vskip 0.2in
\begin{center}
\includegraphics[width=12cm]{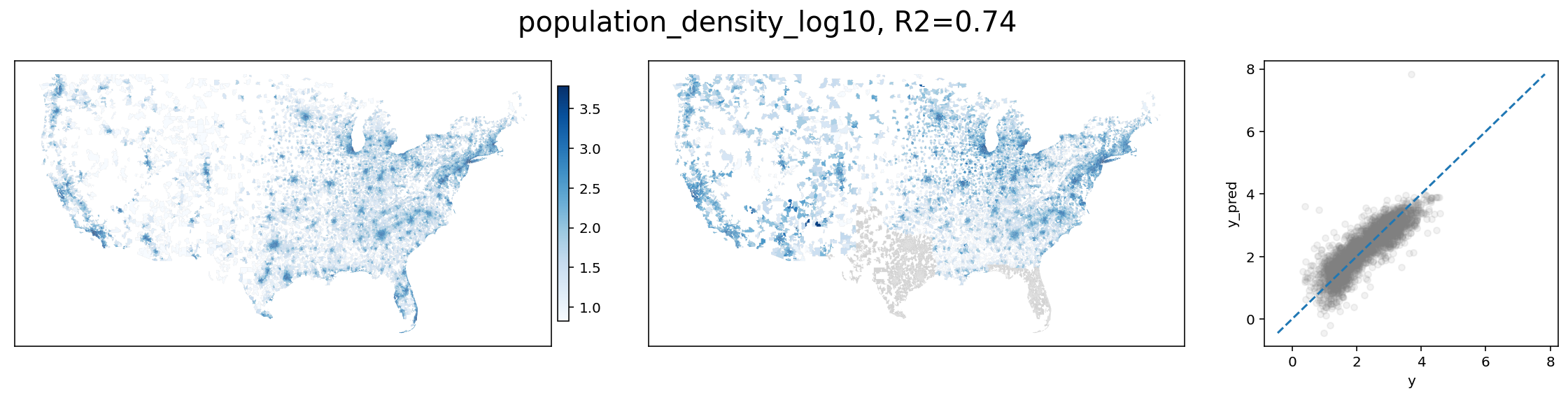}
\includegraphics[width=12cm]{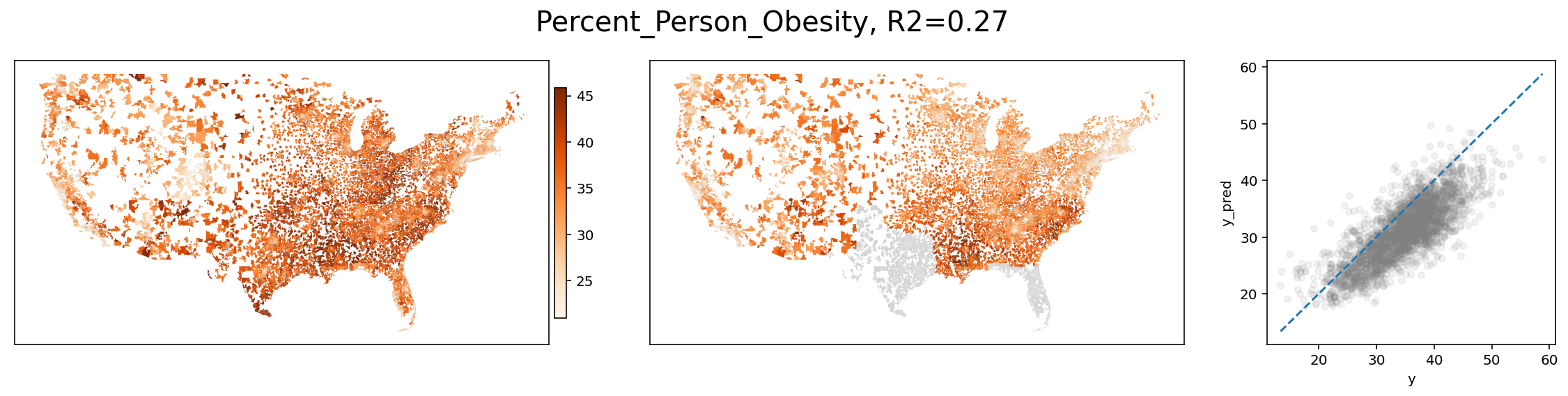}
\includegraphics[width=12cm]{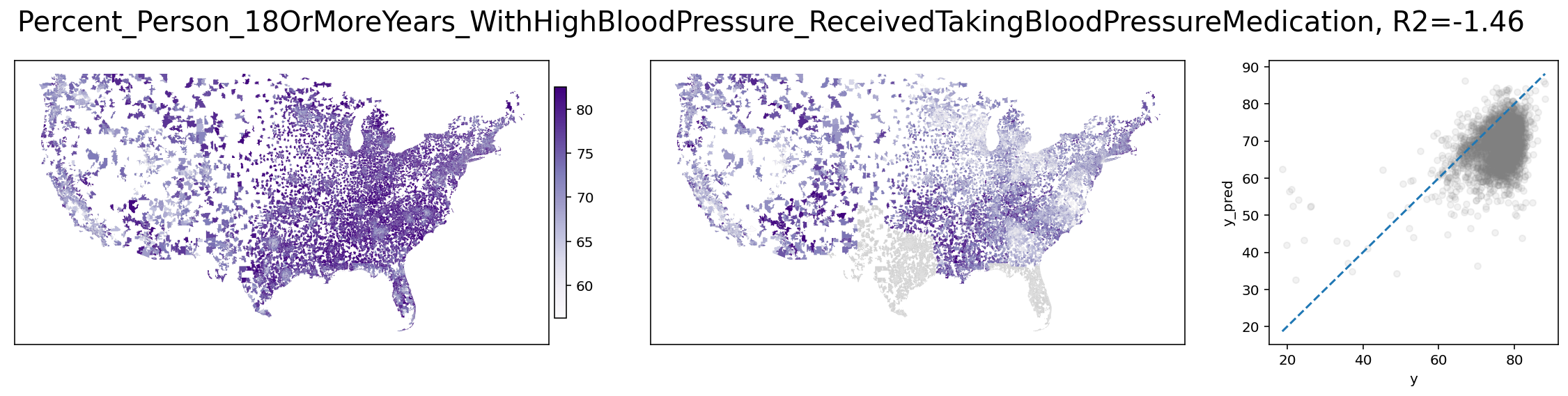}
\caption{Actual, predicted, and validation scatter plot of best, median and worst performing tasks when extrapolating data from Texas and Florida. The left column shows the actual values, the middle column shows the predicted values from the grayed out training points.}
\label{extrapolation}
\end{center}
\vskip -0.2in
\end{figure*}

We also wanted to see if the search features could be used to extrapolate data over large and distant regions. For each task, we trained a model using only search features and labels from two of the southern-most states, Texas and Florida, and tested it on the rest of the study region. The results were mixed, the R\textsuperscript{2} values ranged from -1.47 for predicting blood pressure medication prevalence at the worst, to 0.74 for predicting population density at the best. The median R\textsuperscript{2} across all 27 tasks was 0.27, for predicting obesity prevalence. These three tasks are visualized in Figure \ref{extrapolation}. These experiments show that extrapolation from only a couple of states is unreliable and further study is needed to determine why it works for some tasks and not others.

\subsection{Super-Resolution Tasks}

\begin{figure*}[h!]
\vskip 0.2in
\begin{center}
\includegraphics[width=12cm]{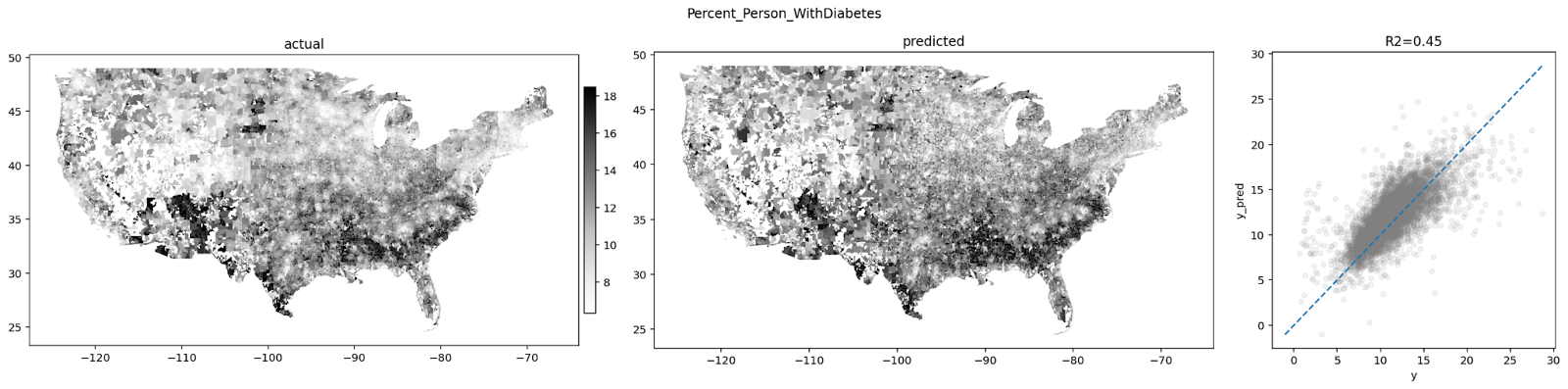}
\includegraphics[width=12cm]{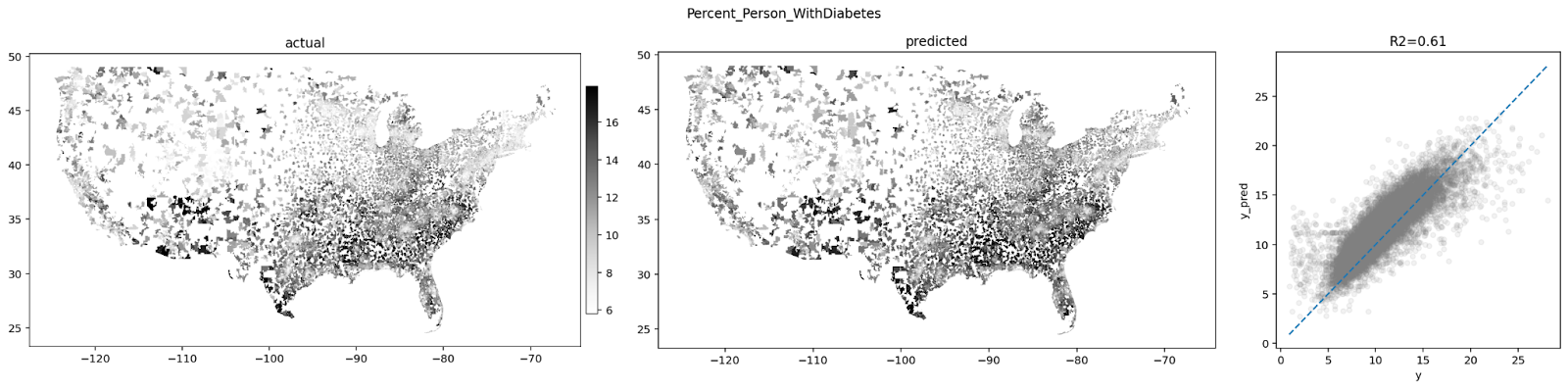}
\includegraphics[width=12cm]{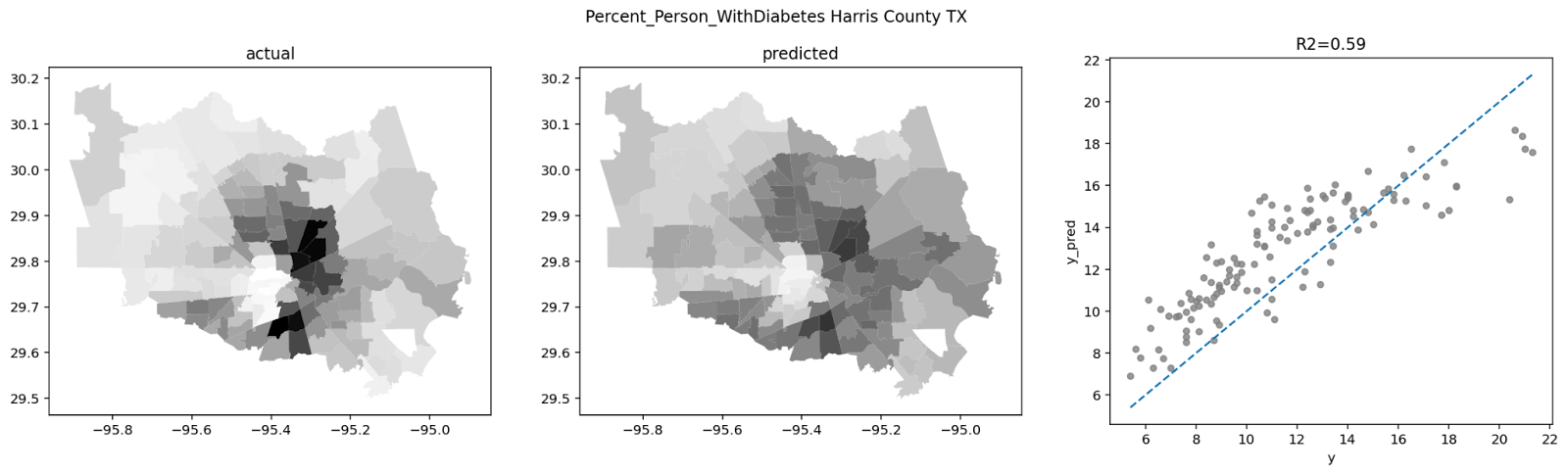}
\caption{Super-resolution results for diabetes prevalence in all zip codes, high population zip codes, and zip codes within Harris County, TX.}
\label{super}
\end{center}
\vskip -0.2in
\end{figure*}

We experimented with training only on county level data and predicting for all zip codes. The search features are aggregated to the county level by taking an unweighted mean of the zip code features within the county.  The county-level labels are directly fetched from Data Commons \cite{datacommons2024}, which includes all 21 health variables as well as median income and median home value.  For each prediction task, a linear regression model is trained on county-level search features and labels. At inference time, zip code level search features are provided to the model for predicting zip code level labels. The mean R\textsuperscript{2} of all variables is 0.64 where the population size is over 3000, and 0.47 over regions. See Figure \ref{super} for a qualitative visualization of our super-resolution results predicting diabetes in all zip codes, zip codes with high populations, and zip codes within Harris County, TX.

\section{Conclusion}

We have demonstrated that a general set of relative search frequencies at the community level can serve as useful foundation features for downstream geospatial modeling tasks such as imputation, extrapolation, and super-resolution. These tasks encompass predicting a wide range of health, socio-economic, and environmental variables across the United States. Notably, our approach does not depend on precise temporal alignment or curation of domain-specific keywords. While this work focused on using simple linear models to assess the ability of these search signatures to encode varied and useful information, in future work we will explore how deep learning models could make use of these features, and assess the incremental improvements in performance from combining them with other features such as aggregated and anonymized mobility patterns, geographic and climate information, and a range of population demographics data. We also hope to extend this analysis globally and to investigate whether a time series of search signatures could be used for effective nowcasting. 

\section*{Broader Impact Statement}

This paper presents work whose goal is to advance the field of 
Machine Learning. There are many potential societal consequences 
of our work, none which we feel must be specifically highlighted here.

\section*{Reproducibility Statement}
The search features dataset described in this paper is currently not publicly available. The Google Trends website may be used to construct similar features at the city-level. All other datasets are publicly available as detailed in the \hyperref[sec:benchmark]{Benchmark Construction section}. The models described can be applied using open source libraries.

\bibliography{dmlr_paper}
\bibliographystyle{icml2024}

\end{document}